\documentclass[runningheads]{llncs}
\usepackage{graphicx}

\usepackage{tikz}
\usepackage{comment}
\usepackage{amsmath,amssymb} 
\usepackage{color}

\usepackage{amsmath}
\usepackage{amssymb}
\usepackage{booktabs}
\usepackage[export]{adjustbox}
\usepackage{xcolor}
\usepackage{bbm}

\usepackage[accsupp]{axessibility}  

\usepackage{hyperref}

\begin{document}
\pagestyle{headings}
\mainmatter
\def\ECCVSubNumber{1512}  

\title{GAMa: Cross-view Video Geo-localization} 

\titlerunning{GAMa: Cross-view Video Geo-localization}
%
\author{Shruti Vyas\and
Chen Chen \and
Mubarak Shah}
\authorrunning{S. Vyas et al.}
%
\institute{Center for Research in Computer Vision, University of Central Florida, USA\\
\email{\{shruti,chen.chen,shah\}@crcv.ucf.edu}}

\maketitle

\begin{abstract}

The existing work in cross-view geo-localization is based on \textit{images} where a ground panorama is matched to an aerial image. In this work, we focus on \textbf{ground videos} instead of images which provides additional contextual cues which are important for this task. There are no existing datasets for this problem, therefore we propose \textbf{GAMa dataset}, a large-scale dataset with ground videos and corresponding aerial images. We also propose a novel approach to solve this problem. At \textbf{clip-level}, a short video clip is matched with corresponding aerial image and is later used to get \textbf{video-level} geo-localization of a long video. Moreover, we propose a hierarchical approach to further improve the clip-level geo-localization. It is a challenging dataset, unaligned and limited field of view, and our proposed method achieves a Top-1 recall rate of 19.4\% and 45.1\% @1.0mile. Code and dataset are available at following
\href{https://github.com/svyas23/GAMa}{link}.

\end{abstract}

\section{Introduction}

Video geo-localization is a challenging problem with many applications such as navigation, autonomous driving, and robotics \cite{li2019cross,tian2021uav,grigorescu2020survey}. The problem to estimate geo-location of the source of a ground video is also faced by first respondents now and then. To solve this problem, there are two main approaches; same-view geo-localization \cite{sarlin2020superglue,regmi2021video,chaabane2021end} and cross-view geo-localization \cite{liu2019lending,zhu2021vigor,regmi2019cross}. In same-view geo-localization, the query ground image is matched with a street view image from the reference set or gallery. Research in videos is limited to same-view geo-localization where a frame-by-frame approach is followed. This approach relies on availability of ground view images for all the locations which may not be possible always.

In such scenarios, cross-view geo-localization is more useful where the query image is matched with the corresponding aerial or satellite image. However, cross-view geo-localization is a much difficult problem since there is a large domain shift between ground and aerial view. 
The limited field-of-view in the ground view makes this even harder and it is sometimes difficult even for humans to identify the correct location of a given image or video.

\begin{figure}[t!]
\begin{center}
\includegraphics[width=0.55\linewidth]{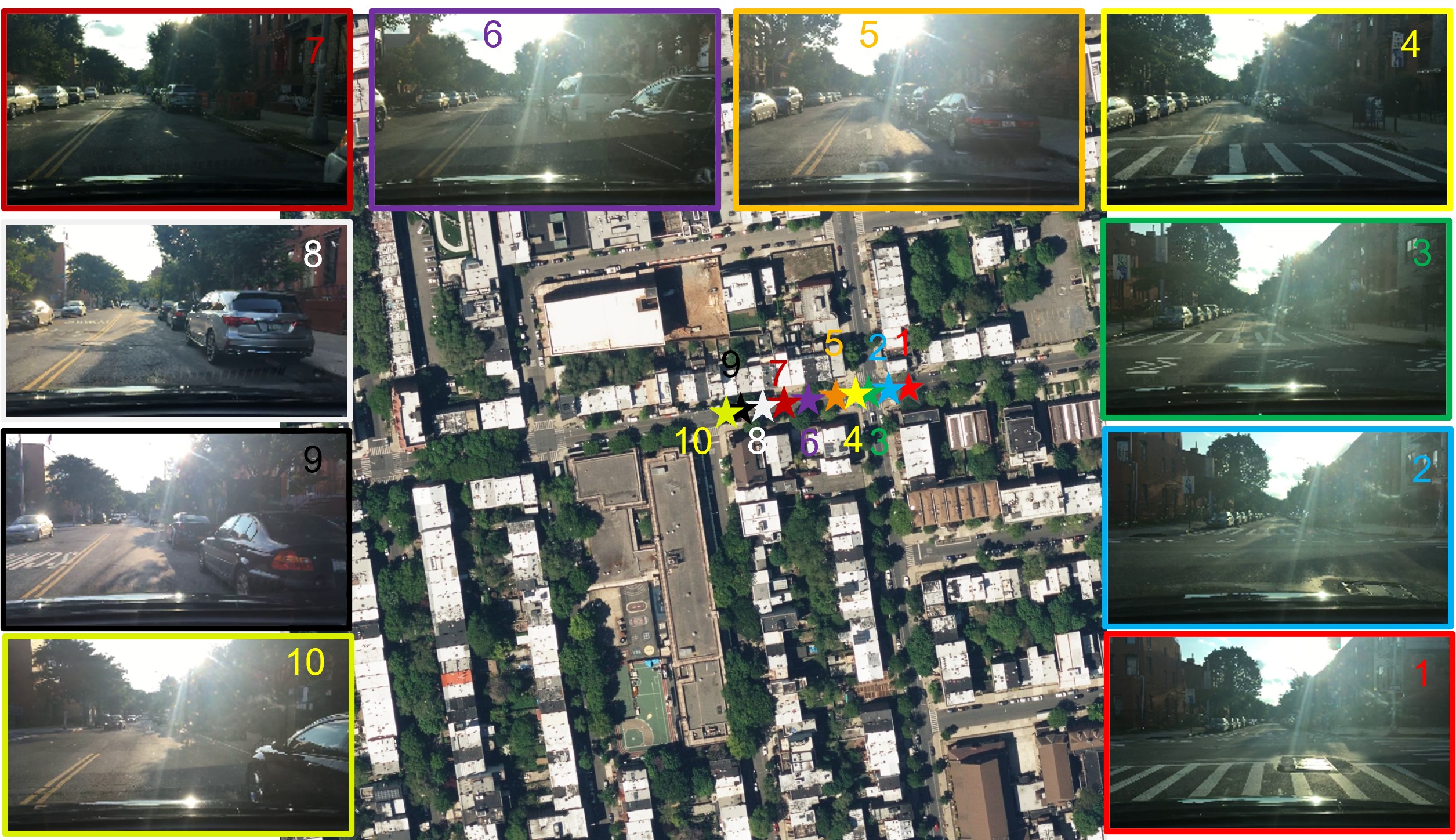}
\enskip
\includegraphics[width=0.3\linewidth]{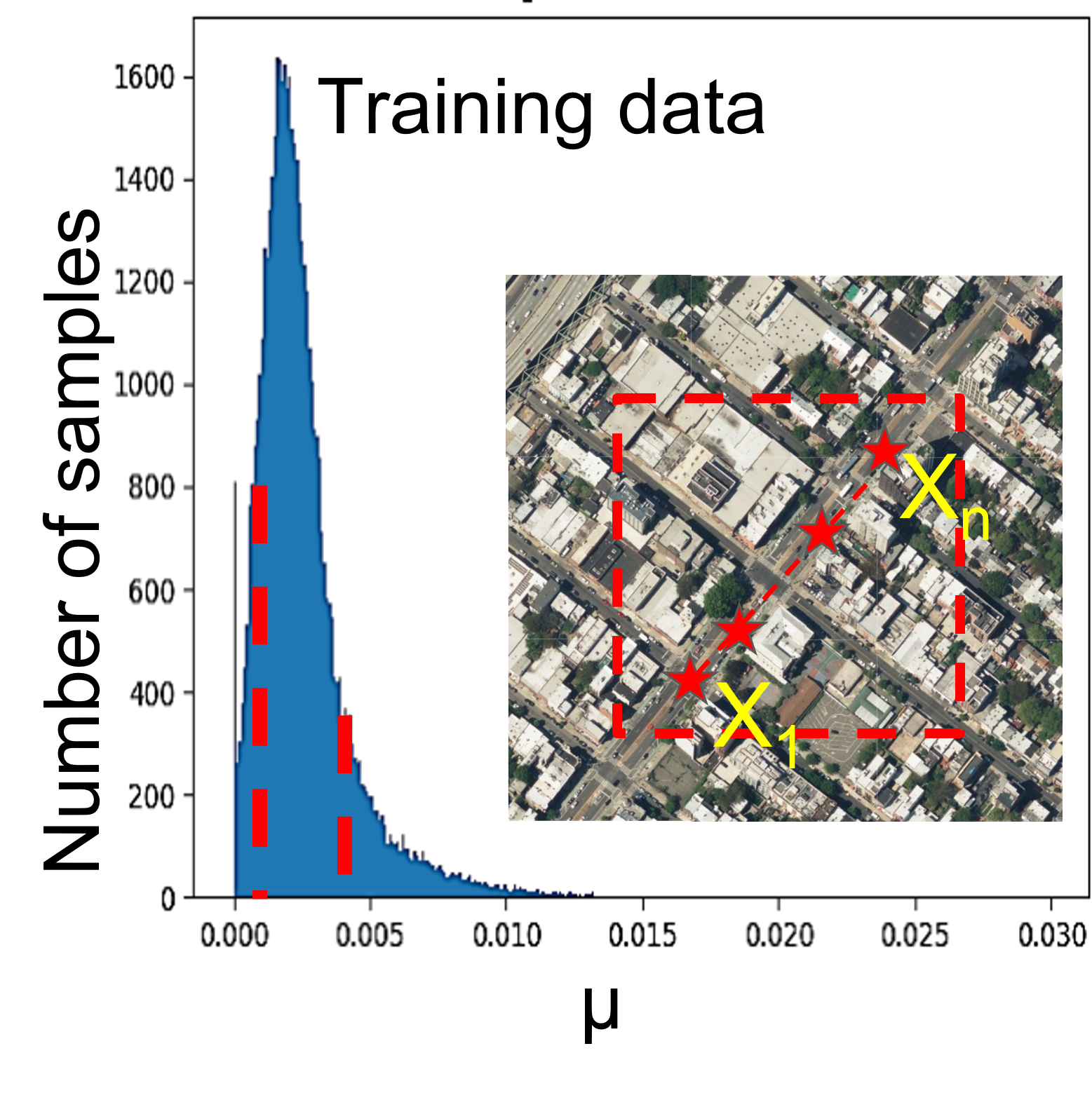}
\end{center}
\caption{\textit{Left:} Sample ground video and aerial image pair from the proposed GAMa dataset. The stars represent location of video frames every sec. and mark the trajectory of the video; \textit{Right:} An example of larger aerial region for a video where stars represent the GPS points labeled every second. The GPS points are marked as $X_{1}$, $X_{2}$, .. $X_{n}$. Range $\mu$ is derived from these GPS points and we also show a histogram of the number of samples at a given value of $\mu$, in training set. The red dotted lines show the lower and and upper threshold (i.e. 0.001 and 0.004) for selecting the videos}
\label{fig:dataset_sample}
\vspace{-18pt}
\end{figure}

The existing works in cross-view geo-localization follow an image based approach where a ground image is matched with an aerial image \cite{regmi2019cross,wang2021each,toker2021coming}. 
In such an approach, the contextual information available with the video is lost. We focus on the geo-localization of ground videos instead of images to utilize the context, i.e. how the view in one frame is related/located w.r.t. another frame.
To the best of our knowledge there are no existing datasets which are publicly available and can be used for this problem. Thus, we have collected a new dataset, named as \textit{GAMa (\textbf{G}round-video to \textbf{A}erial-image \textbf{Ma}tching)}, which contains ground videos with GPS labels and corresponding aerial images. It consists of $\sim$ 1.9M aerial images and 51K ground videos where each video is around 40 seconds long. An example video with representative frames and corresponding aerial image is shown in Figure \ref{fig:dataset_sample}.
We will make this dataset publicly available for further research.

We propose GAMa-Net as a benchmark method to solve this problem at clip level where we match every 0.5 second (short clip) from a long video with the corresponding aerial view. 
A frame-by-frame approach can also be used for video geo-localization, where a 2D convolutional network is used to get the spatial features frame-by-frame. However, following a frame-by-frame approach ignores the rich contextual information available in a video. Considering this, we propose a 3D-convolution based approach for learning the ground level features from a video. 
The proposed approach is trained using an \textit{image-video contrastive loss} which tries to match the ground view with the aerial view. This matching provides a \textit{clip-level} geo-localization. 


Next, we propose a \textit{hierarchical approach} which helps in improving the clip-level geo-localization performance while providing a \textit{video-level} geo-localization with the help of clip-level predictions. It takes the set of aerial images corresponding to the clips of a long video, and matches them against a larger geographical area. Therefore, it makes use of the contextual information available with the sequence of clips corresponding to a longer video. 

We evaluate the proposed approach on GAMa dataset and demonstrate its effectiveness for clip/video geo-localization. We provide an analysis and propose a set of baselines to benchmark the dataset. We make the following contributions:
\begin{itemize}
\setlength\itemsep{-.1em}
\item A novel problem formulation \textit{i.e.} cross-view \textit{video} geo-localization and a large-scale dataset, \textit{GAMa}, with ground videos and corresponding aerial images. This is the \textit{first} video dataset for this problem to the best of our knowledge. 
\item We propose \textit{GAMa-Net}, which performs \textit{cross-view video geo-localization} at \textit{clip-level} by matching a ground video with aerial images using an \textit{image-video contrastive} loss.
\item We also propose a novel \textit{hierarchical approach} which provides \textit{video-level} geo-localization and utilizes aerial images at different \textit{scales} to improve the clip-level geo-localization. 
\end{itemize}

\section{Related works}
\label{sec:related}
Traditional features of classical computer vision were earlier used for image matching in geo-localization \cite{senlet2012satellite,zamir2010accurate}. As deep learning has proven successful in feature learning most of the recent studies have adopted CNN based approach for learning discriminative features for image matching \cite{tian2017cross}. The problem of image or video geo-localization is solved using either the same view, which is mostly the ground view, or cross-view. \textbf{Same-view Geo-localization} makes use of the large collections of geo-tagged images available online \cite{radenovic2018fine,zemene2018large,sarlin2020superglue,regmi2021video,chaabane2021end,arandjelovic2016netvlad}. The problem is approached with the assumption that there is a reference dataset consisting of geo-tagged ground images and there is an image corresponding to each query image. The problem is then solved as image retrieval of the matching reference to determine the location of the query. There is some research in video geo-localization as well which is solved at frame level which is followed by trajectory smoothing \cite{hakeem2006estimating,regmi2021video,chaabane2021end}. However, a more complete coverage by overhead reference data such as satellite/aerial imagery has spurred a growing interest in cross-view geo-localization.

\textbf{Cross-view Geo-localization.} Most of the recent work adopt CNN based approaches. Several studies have explored CNN architectures for matching ground-level query images to overhead satellite images \cite{vo2016localizing,toker2021coming,zhu2021geographic,lin2015learning,hu2018cvm,wang2021each}. Triplet loss is mostly used optimization function in these studies \cite{hu2018cvm,regmi2019cross}. Certain studies however report better results with contrastive loss\cite{radenovic2018fine}. Field-of-view (FOV) also plays an important role in deciding the recall rate and ground panorama is highly accurate as compared to limited FOV \cite{hu2018cvm,zhu2021vigor}. Similarly, videos also contain more visual information as seen through the trajectory of the camera and can be expected to provide more accurate geo-localization as compared to images or frames with similar FOV.

Current works on cross-view geolocalization follow image based approach since the existing datasets only contain image pairs for ground and aerial view  \cite{tian2017cross,regmi2019cross,hu2020image,yang2021cross,rodrigues2021these}. However, some papers do report testing their model on videos using a frame-by-frame approach \cite{hu2020image}. Most popular datasets, Cross-View USA (CVUSA) dataset \cite{workman2015wide}, CVACT \cite{liu2019lending}, and Vo et al. \cite{vo2016localizing} contain ground panorama aligned with corresponding aerial image. Recent publications have shown very high recall rate on these datasets while using panoramas however these values are quite low when using limited FOV and unaligned images, i.e. top@1 recall of upto 14\%  \cite{shi2020looking,yang2021cross}. VIGOR \cite{zhu2021vigor} dataset also contains panorama however being unaligned it is more realistic. 
All these datasets use ground panorama which is not realistic from video geo-localization, as videos have limited FOV, neither do they have time series data required for such training. It is possible to get unaligned images and limited FOV from these datasets however, there is no existing dataset with ground videos and aerial image pairs to solve \textbf{cross-view} geolocalization in videos and the proposed dataset addresses this gap.

Cross-view is also used for fine geo-localization of UAVs or robots. Camera feed (also frame-by-frame) and a known small region/map, of about a mile is used to find a more exact location in the given map \cite{hosseinpoor2016pricise}. Sometimes a prior is given to estimate the vehicle pose \cite{kim2017satellite}. However, our focus is on coarse geo-localization where the gallery spans over multiple cities. 


A frame-by-frame geolocalization of videos is also possible with the proposed dataset where 2D convolutional networks are used to extract the spatial features. However, while following a frame-by-frame approach the contextual information as available from a video is ignored. Fusion of features obtained from 2D-CNN is also possible however it is more memory intensive as compared to a 3D-CNN network. Considering these limitations and challenges, we propose a videos based cross-view geo-localization. 

\setlength{\tabcolsep}{4pt}
\begin{table}[t!]
\begin{center}
\caption{Statistics of the proposed GAMa dataset. Note: Large aerial images are at 1792X1792 resolution whereas small aerial images or tiles are at 256x256 resolution}
\label{table:dataset}
\begin{tabular}{lcccc}
\hline\noalign{\smallskip}
\textbf{Parameter} & \textbf{Train} & \textbf{Test} & \textbf{Train(day)} & \textbf{Test(day)} \\
\noalign{\smallskip}
\hline
\noalign{\smallskip}
Videos & 45029 & 6506 & 21144 & 3103 \\ 
Large aerial images & 45029 & 6506 & 21144 & 3103 \\
\midrule
Clips & 1.68M & 243k & 790k & 116k \\ 
CN small aerial images & 1.68M & 243k & 790k & 116k \\ 
UCN small aerial images & 2.21M & 319k & 1.04M & 152k \\
\hline
\end{tabular}
\end{center}
\end{table}
\setlength{\tabcolsep}{1.4pt}

\section{GAMa dataset}

The proposed GAMa (\textbf{G}round-video to \textbf{A}erial-image \textbf{Ma}tching) dataset comprises of select videos from BDD100k \cite{yu2020bdd100k} and aerial images from apple maps. 

\textit{Ground video selection:}
Most of the videos in BDD100k dataset are 40 sec long and usually have GPS label every second. 
The selection of videos from dataset was based on the range of latitude and longitude for a given video where we use a range parameter $\mu$. We label the GPS points at $n^{th}$ second as $X_{n}$ ($lat_{n}$, $long_{n}$), where the corresponding latitude is $lat_{n}$ and longitude is $long_{n}$ (Figure \ref{fig:dataset_sample}\textit{R}, Aerial image). 
Thus, for the whole video we have GPS points as $X_{1}$($lat_{1}$, $long_{1}$), $X_{2}$($lat_{2}$, $long_{2}$),.., $X_{n}$($lat_{n}$, $long_{n}$). If max latitude = $lat_{k}$ and min latitude = $lat_{l}$, then Latitude range = $lat_{k}$ – $lat_{l}$. Also, if max longitude = $long_{p}$ and min longitude = $long_{q}$, then Longitude range = $long_{p}$ – $long_{q}$. 
Range, $\mu$ = max(Latitude range, Longitude range)
In order to eliminate stationary and very fast videos, we select videos with $\mu$ from 0.001 to 0.004. Figure \ref{fig:dataset_sample}\textit{R}. 
shows the distribution of training videos with range, $\mu$. The distribution was similar for training and test sets. This selection left us with 46596 training and 6728 testing videos which were further screened based on the availability of GPS labels.

\textit{Aerial images:}
For the selected videos, aerial images are downloaded as tiles from Apple maps at 19x zoom \cite{apple_maps}. The dataset comprises of one large aerial image (1792x1792) corresponding to each video of around 40 sec. and 49 uncentered small aerial images (256x256) for these large aerial regions. Table \ref{table:dataset} summarizes the dataset statistics. Since, most of the videos have a GPS label every second we divide the videos into smaller clips of 1 sec. each and for each clip we have an aerial tile.

In Figure \ref{fig:dataset_sample}, we see an example of a large aerial region, along with the video frames at each second. 

\begin{figure}[t!]
\begin{center}
\includegraphics[width=0.45\linewidth]{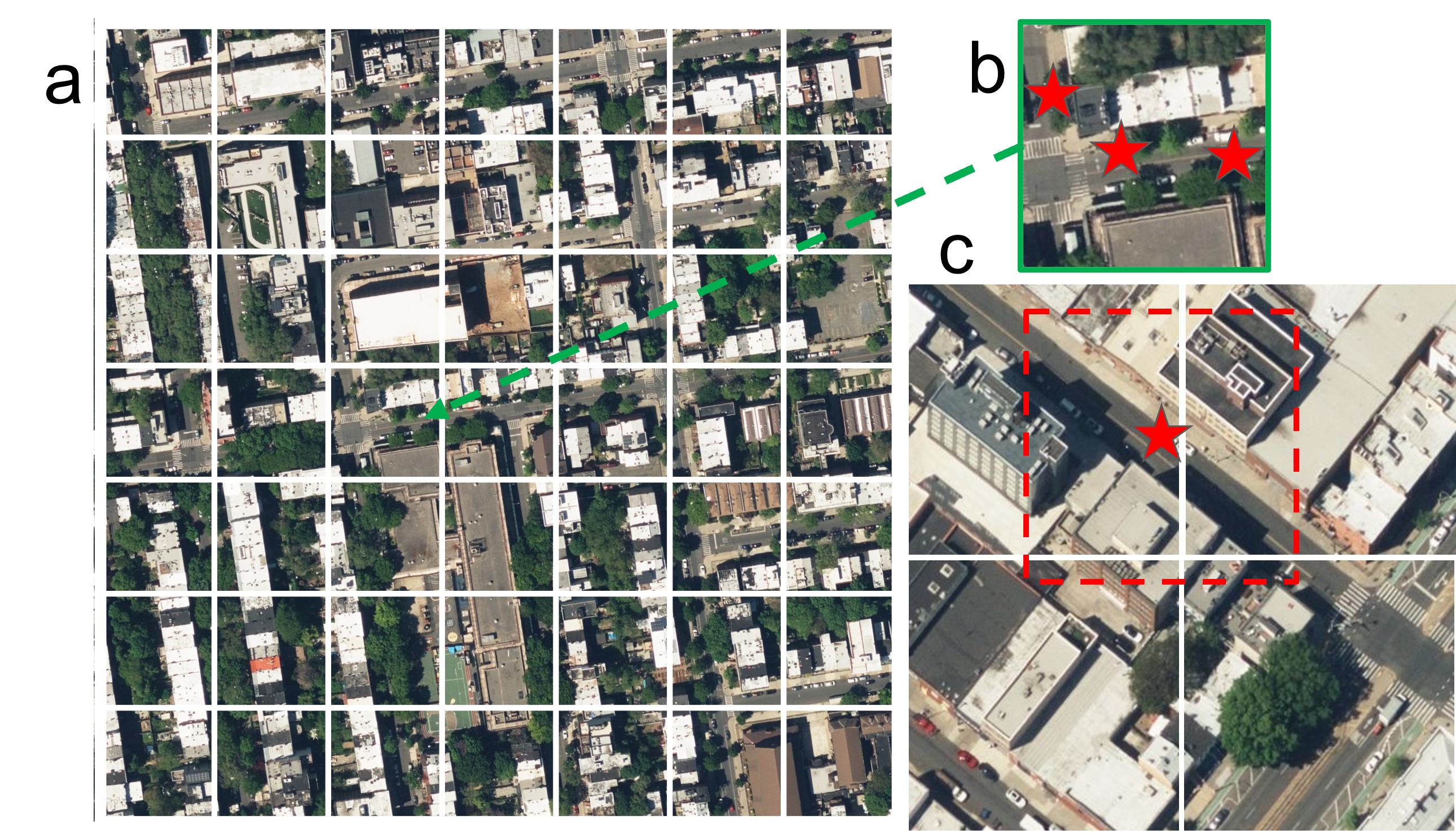}
\enskip \quad \quad
\includegraphics[width=0.42\linewidth]{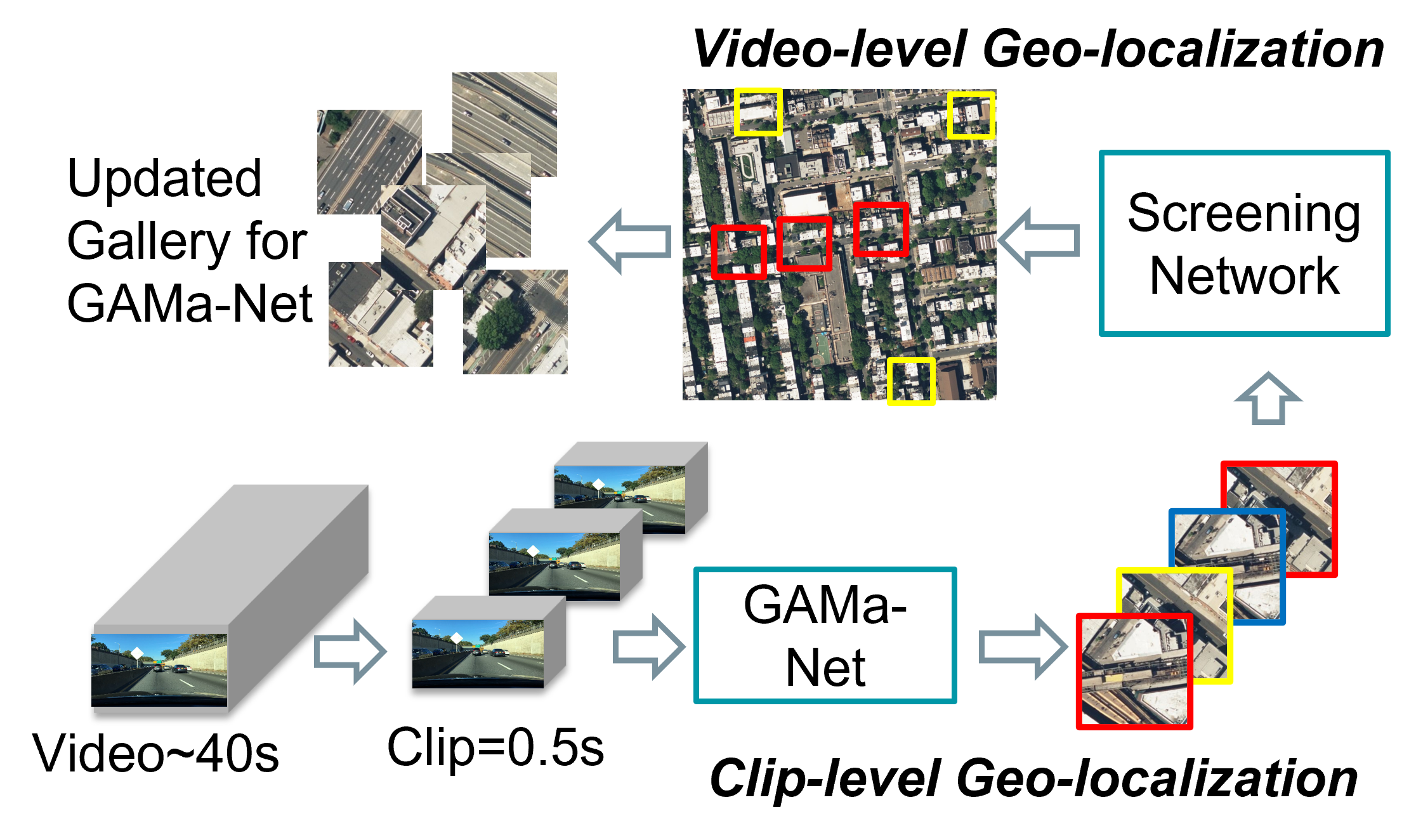}
\end{center}
  \caption{\textit{Left(L)}: Sample aerial images from GAMa dataset. a) A large aerial image matched to a video and division into small uncentered(UCN) aerial images, b) Small UCN aerial image (Zoomed-in) and possible GPS labels for a matching clip, and c) small centered (CN) aerial image as centered around the GPS label; \textit{Right(R)}: An outline of the proposed approach. From a given ground video, clips of 0.5 sec are input to GAMa-Net, one clip at a time is matched to an aerial image. The sequence of aerial images thus obtained from a video is input to the Screening network to retrieve the corresponding large aerial region for video-level geolocalization. Top predictions of larger aerial regions provide the updated gallery for GAMa-Net}
\label{fig:dataset_sample_larger_smaller_centering}
\vspace{-10pt}
\end{figure}

\textit{Aerial image centering:}
In Figure \ref{fig:dataset_sample_larger_smaller_centering}\textit{(L)}a, we show an example of a large aerial image. We have a centered (CN) and an uncentered (UCN) set of small aerial images, Table \ref{table:dataset}. UCN aerial images are obtained by dividing the large aerial image into smaller tiles. The GPS label in these UCN smaller aerial images can be anywhere besides the center, Figure \ref{fig:dataset_sample_larger_smaller_centering}\textit{(L)}b. There are three labels in the figure and for each of these GPS points the same tile will be considered as the ground truth. For making a centered (CN) set, as shown in Figure \ref{fig:dataset_sample_larger_smaller_centering}\textit{(L)}c we take a crop centered around the corresponding GPS point however it has lead to overlap among the aerial images since in some videos we have a distribution where GPS points are nearby. In the dataset, we still have a one on one correspondence among aerial images and short ground clips. The overlap among the aerial images however increases the difficulty level for top-1 retrieval thus we evaluate with distance threshold.

The dataset covers multiple cities, and the ground videos are distributed all over the US and also from middle eastern region. However, most of the videos are from four US cities; New York, Berkeley, San Francsico, and Bay area. They show different weather conditions, including sunny, overcast, and rainy, as well as different times of the day including day and night. 
 
There is occlusion in videos and shadows of the skyscrapers in aerial images. Limited FOV in videos and all stated characteristics bring it closer to a realistic scenario however also makes it a difficult dataset for geolocalization. 
In Table \ref{table:dataset_comparison}, we show a comparison with existing datasets for cross-view geolocalization.

\setlength{\tabcolsep}{4pt}
\begin{table}[t!]
\begin{center}
\caption{Comparison of GAMa dataset with existing cross-view geo-localization datasets. Please, note that the previous datasets do not contain ground videos}
\label{table:dataset_comparison}
\begin{tabular}{lcccc}
\hline\noalign{\smallskip}
& Vo\cite{vo2016localizing} & CVACT\cite{liu2019lending} & CVUSA\cite{workman2015wide} & GAMa (proposed) \\
\noalign{\smallskip}
\hline
\noalign{\smallskip}
Ground videos & no & no & no & 51535 \\ 
Panorama & ~450,000 & 128,334 & 44,416 & no \\
Aerial images & ~450,000 & 128,334 & 44,416 & 1.92M \\
Aerial img resolution & - & 1200x1200 & 750x750 & 256x256 \& 1792x1792\\
Multiple cities & yes & no & yes & yes \\
\hline
\end{tabular}
\end{center}
\end{table}
\setlength{\tabcolsep}{1.4pt}

\section{Method}
An overview of the proposed approach which works on clip and video level is show in Figure \ref{fig:dataset_sample_larger_smaller_centering}\textit{R}. 
Ground-video to Aerial-image Matching Network (GAMa-Net) learns features for ground view clips and aerial images; and bring the matching pair closer in the feature space by applying a contrastive loss. This provides a clip-level geolocalization for a long video. In addition, we propose an hierarchical approach, where we do video-level geolocalization and use it to update the gallery of aerial images by selecting top matched large aerial regions (Figure \ref{fig:dataset_sample_larger_smaller_centering}\textit{R}). 
The reduced gallery helps improve the clip-level geolocalization by screening out some of the visually similar however incorrect aerial images. 

\subsection{GAMa-Net: \textit{Clip-level Geo-localization}}

\begin{figure}[t!]
\begin{center}
\includegraphics[width=0.9\linewidth]{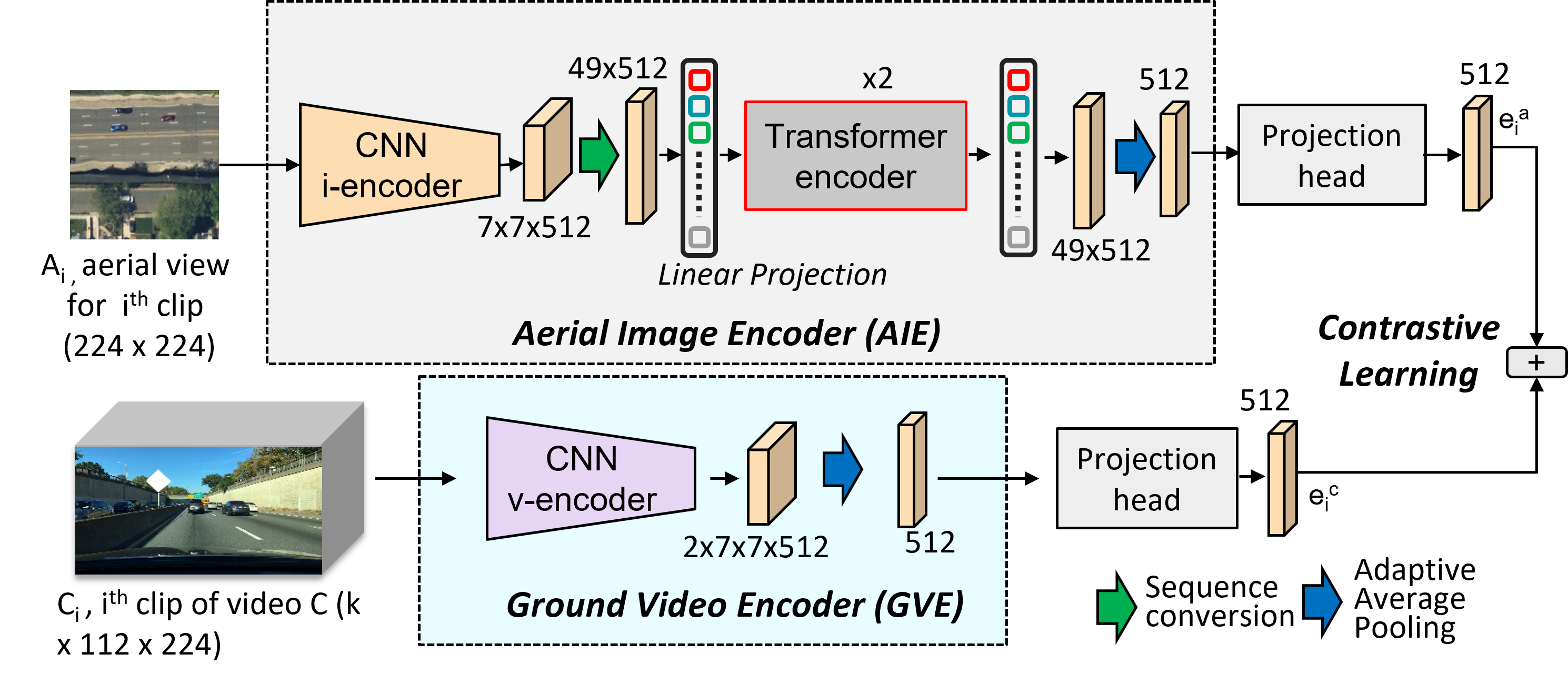}
\end{center}
\caption{Network diagram for \textbf{GAMa-Net} proposed for clip-level geolocalization. We use 3D-CNN as our base network for learning features from a ground clip, around 0.5sec. Similarly, for aerial image features, we use a 2D CNN backbone. Since only some part of aerial images are covered by the video feed, using a transformer encoder improves the learning. Number of frames, k=8 with skip rate=1}
\label{fig:network_sat_attention}
\vspace{-10pt}
\end{figure}

The proposed network takes as input a short clip from a ground video and matches it with corresponding aerial image. An overview of the proposed network is shown in Figure \ref{fig:network_sat_attention}. 

\textit{\textbf{Visual encoders}} In GAMa-Net, we have a video encoder i.e. Ground Video Encoder (GVE) to get features from ground video frames and an image encoder for aerial image features i.e. Aerial Image Encoder (AIE). GVE uses 3D-ResNet18 as backbone and a two layer transformer encoder. Given a ground video C, GVE provides features for a 8 frame clip $C_{i}$ at the $i^{th}$ second of the video. There is a skip-rate of one frame so we are covering around half a second in the clip, $C_{i}$. AIE on the other hand uses 2D-ResNet18 as the backbone. It takes the corresponding aerial image, $A_{i}$ as the input and learns the features to match with that of the ground video.

\textit{\textbf{Transformer encoding:}} In the ground video all the visual features are not of equal importance for matching with the aerial view. This is true for aerial images as well and few features are more important when matching with ground videos. For example, the top of a building as seen from aerial images is not visible from a ground video and hence cannot be used to match the pair. To address this, we experimented with multi-headed self attention. We input the features obtained from convolutional networks to a transformer encoder framework which comprises of 4 heads and 2 layers, and uses positional encoding.
A small neural network i.e. projection head is used to map the representations to the space where contrastive loss is applied. We use a MLP with one hidden layer to obtain the ground video and aerial image feature vectors ${e^c}_{i}$ and ${e^a}_{i}$, respectively.
 
\textit{\textbf{Image-video contrastive loss:}} We utilize contrastive loss formulation, base on NT-Xent \cite{chen2020simple,NEURIPS2019_9015,sohn2016improved}, to train our network. For a given ground video the corresponding aerial image is considered a positive sample and rest of the samples are considered as negatives. This is a image-video contrastive loss applied on features from two different visual modalities i.e. ground videos and aerial images. 
In the loss formulation, the focus is on reducing the distance between the positive pair. We have a total of $2(N)$ data points in any mini-batch with $N$ examples. The image-video contrastive loss for a pair of positive examples is defined as,
\begin{align}
\label{eq_contrastive}
l_{c,a} = -\log\frac{\exp(\mathrm{sim}({e^c}_{i},{e^a}_{i})/\tau)}{\sum^{N}_{k=1}\mathbbm{1}_{[k \neq i]}\exp(\mathrm{sim}({e^c}_{i}, {e^c}_{k})/\tau) + \sum^{N}_{k=1}\mathbbm{1}_{[k \neq i]}\exp(\mathrm{sim}({e^c}_{i}, {e^a}_{k})/\tau)},
\end{align}

where ${e^c}_{i}$ and ${e^a}_{i}$ is a positive pair, ${e^c}_{i}$ and (${e^c}_{k}$ or ${e^a}_{k}$)  are negative pairs, $\mathbbm{1}_{[k \neq i]} \in \{0,1\}$ is an indicator function with value 1 if $k \neq i$, $\tau$ is a temperature parameter, $\mathrm{sim}$ is the cosine similarity between a pair of features. The final loss is computed for all the positive pairs, both $(e^c,e^a)$ and $(e^a,e^c)$.

\subsection{Hierarchical approach} 

In this approach, we introduce video-level geolocalization which also helps in reducing the search space for GAMa-Net (Figure \ref{fig:dataset_sample_larger_smaller_centering}\textit{R}). 
The clips of a given video are temporally related and provide contextual information to help improve the geolocalization. Similarly, while looking at this problem from aerial image view point, the sequence of aerial images corresponding to clips from a given video are also related geographically and some of the correct prediction at clip-level can be used to update the gallery. Using a smaller gallery also reduces the possibility of error in feature matching. 

\textit{\textbf{Approach:}} We have four steps in this approach. In Step-1, we use GAMa-Net which takes one clip (0.5 sec) at a time and matches with an aerial image. Using multiple clips of a video, we get a sequence of aerial images for the whole video, i.e. around 40 small aerial images. In Step-2, we use these predictions of aerial images and match them to the corresponding larger aerial region. We use a screening network to match the features however the features are from the same view i.e aerial view. In Step-3, we use the predictions to reduce the gallery by only keeping top ranked large aerial regions corresponding to a video. These large aerial regions define our new gallery for a given video. In Step-4, we use GAMa-Net i.e. the same network as in Step-1, however geo-localize using the updated gallery.

Visually correct predictions from Step-1 which may not be the ground truth are used to reduce the gallery using this approach. This approach helps improve the clip-level geolocalization since the reduction of the gallery is based on the fact that all the clips of a given video are geolocated nearby. Thus, the aerial images predicted by GAMa-Net can be used to find that large aerial region where all these clips have been captured. In this case, the probability of finding all those visually correct aerial images in the same region is higher when we are searching at the correct geolocation. Hence, it is likely to match with correct large aerial region provided that meaningful and enough information is available. 

\begin{figure}[t!]
\begin{center}
\includegraphics[width=0.6\linewidth]{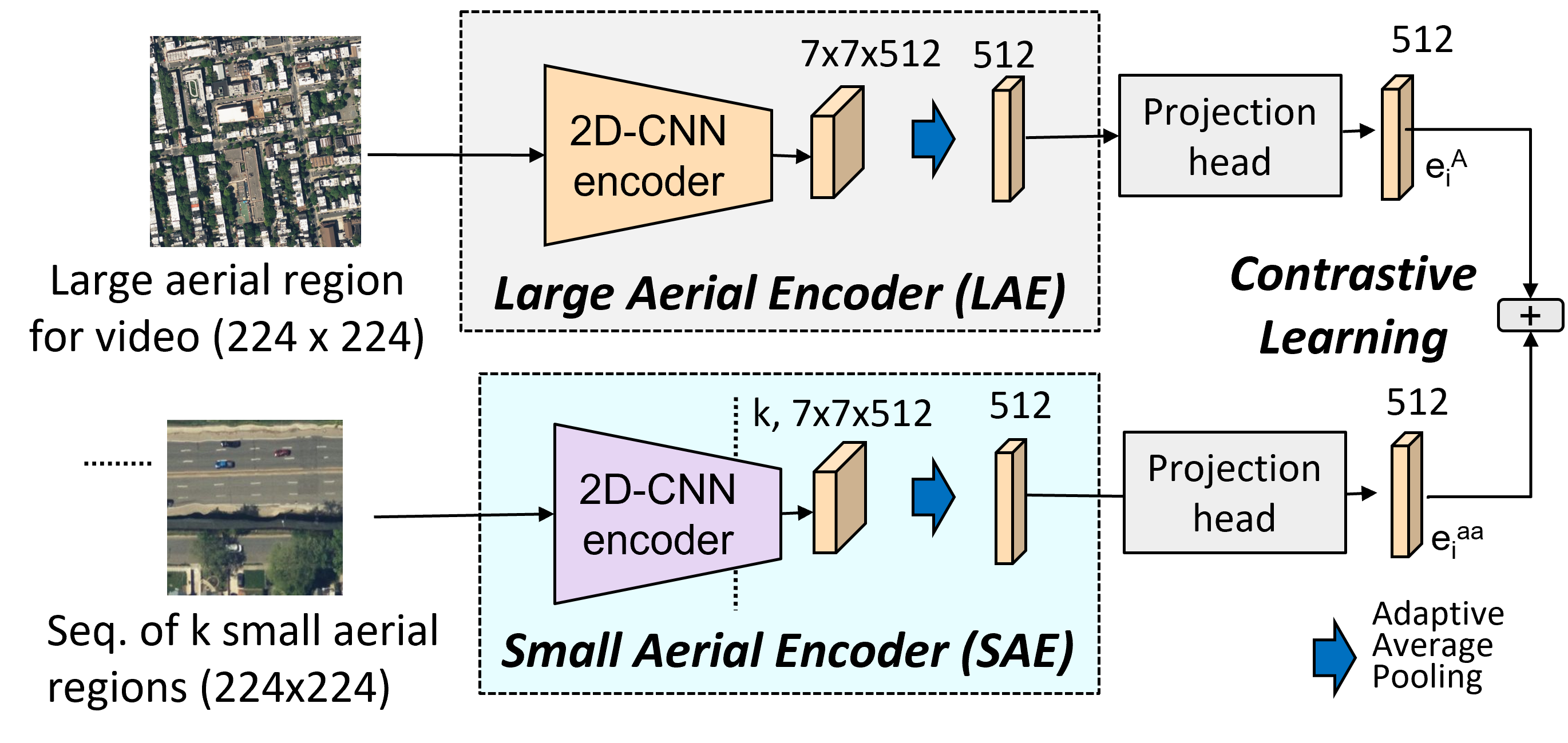}
\enskip
\includegraphics[width=0.35\linewidth]{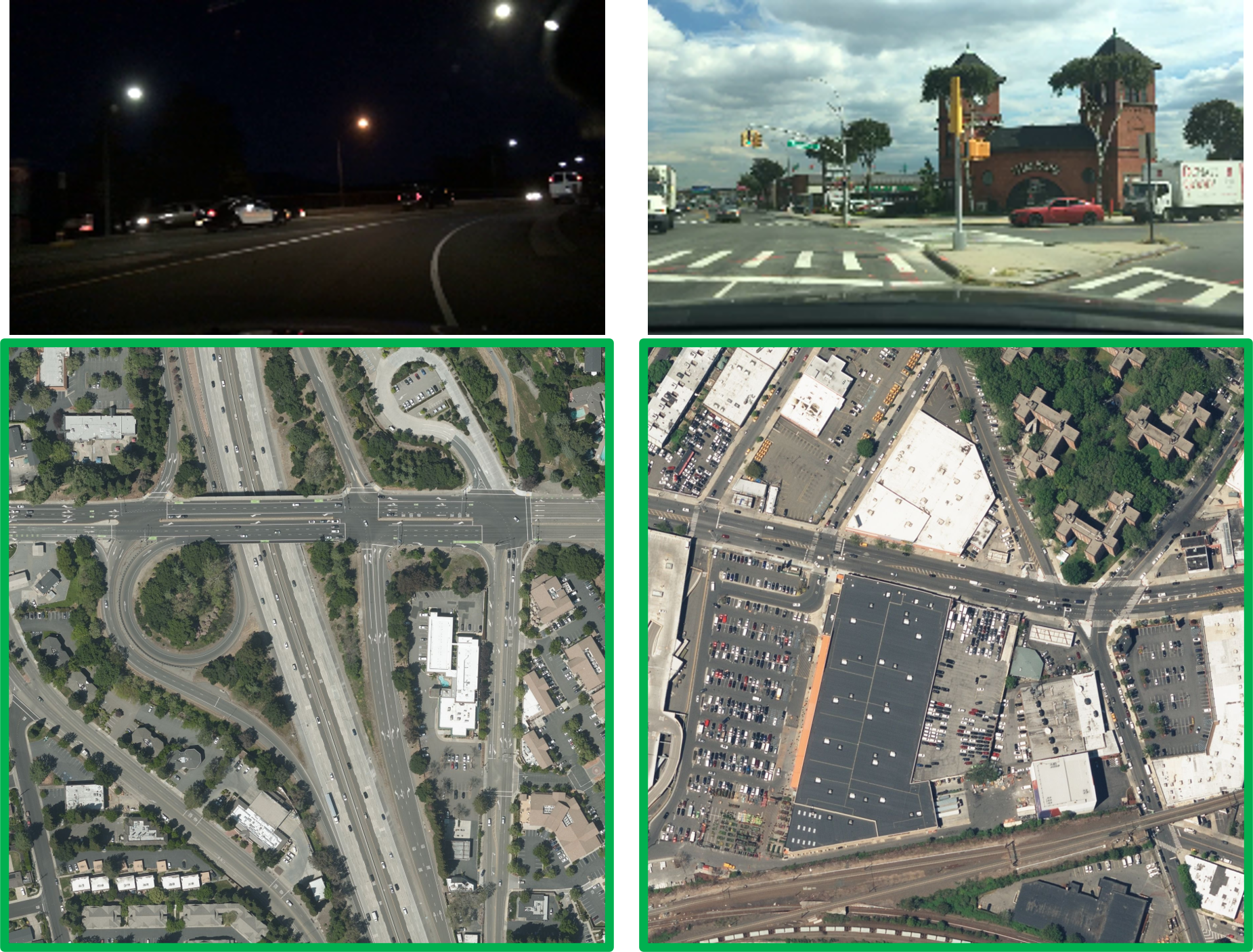}
\end{center}
\caption{\textit{Left(L):} Network diagram for the Screening network used for video-level geo-localization in our proposed Hierarchical approach. \textit{Right(R):} Predictions at video level, where a single frame of video is shown along with the correct matched large aerial region. More examples are in supplementary
}
\label{fig:network_screening}
\vspace{-10pt}
\end{figure}

\textit{\textbf{Screening Network: Video-level geo-localization}} \\We use the screening network to match a sequence of smaller aerial images with the corresponding larger aerial region. The network is similar to GAMa-Net, however the sequence of aerial images is input to Small Aerial Encoder (SAE) with 2D-ResNet18 backbone and the sequence of features are later averaged to obtain a 512D feature vector (Figure \ref{fig:network_screening}\textit{L}). We also experimented with a 3D-ResNet18 backbone which is discussed later. The feature vector thus obtained is matched with the feature vector of corresponding large aerial image from Large Aerial Encoder (LAE). 
We apply contrastive loss similar to Equation \ref{eq_contrastive}. The predictions from this network are used to update the gallery for GAMa-Net. 

\section{Experiments and Results}

\textit{\textbf{Implementation and training details:}} 
We implement our GAMa-Net and screening network using PyTorch \cite{NEURIPS2019_9015} and train using Adam optimizer with a learning rate (lr) of 8e-5. We use a lr scheduler with lr decay rate of 0.1.
The screening network is trained in two step; first with the actual ground truth sequences, and then finetuned with predictions from GAMa-Net. The finetuning step allows the network to adapt to noisy aerial sequences which will be used during inference. We use pre-trained weights from Kinetics-400 for 3D-ResNet18 and ImageNet weights for 2D-ResNet18. The ground videos in the proposed dataset are from different times of the day, for faster training we have used \textbf{only day} videos in our experiments unless stated otherwise.

\textit{\textbf{Evaluation:}} We use top-k recall for clip-level and video-level matching at different values of k. Given a video query, its closest k reference neighbors in the feature space are retrieved as predictions. Similar to image based geolocalization methods we use recall rates at top-1, top-5, top-10, and top 1\%. More details can be found in \cite{vo2016localizing,hu2018cvm,shi2019spatial}. We have UCN and CN sets of aerial images corresponding to each clip. In UCN set there is one-to-one correspondence between the clip and aerial image. The GPS point can be anywhere within the aerial image however in the CN set there is an overlap among the aerial images. To keep the evaluation similar to UCN set it is considered a correct match if the predicted GPS is within the range of 0.05 mile of the correct location.
We also report top-1@t rate where t is a distance threshold to be used for correct prediction. 

\textit{\textbf{Baselines:}} We propose several baselines for comparison. For ground image based baselines
we use the center frame of the clip as an input. Image-CBn, our proposed baseline uses 2D-CNN ResNet18 model to encode the ground video frame with similar contrastive loss formulation as GAMa-Net. We use two different loss functions for video based baselines (Triplet-Bn and CBn); margin triplet loss \cite{BMVC2016_119} and contrastive loss \cite{chen2020simple}. In these baselines, we utilize 2D-CNN ResNet18 for aerial images and 3D-CNN ResNet18 for ground videos. 

\setlength{\tabcolsep}{4pt}
\begin{table}[t!]
\begin{center}
\caption{Comparison with proposed baselines, where CBn-Contrastive Baseline \& Bn-Baseline. Recall rates in parentheses() are with UCN aerial images }
\label{table:method_comparison}
\begin{tabular}{lccccc}
\hline\noalign{\smallskip}
Model & Video & R@1 & R@5 & R@10 & R@1\% \\
\noalign{\smallskip}
\hline
\noalign{\smallskip}
Image-CBn & x & 9.5(1.5) & 18.8(5.5) & 24.6(9.0) & 87.7(50.7) \\
Shi et al.\cite{shi2020looking} & x & 9.6 & 18.1 & 26.6 & 71.9 \\
L2LTR \cite{yang2021cross} & x & 11.7 & 20.8 & 28.2 & 87.1 \\
\midrule
Triplet-Bn & \checkmark & \textless 1(\textless 1) &  \textless 1(1.2) & 1.1(2.4) & 33.1(33.8) \\
CBn & \checkmark &11.5(3.2) & 21.7(10.9) & 27.8(16.7) & 89.2(65.0) \\
\textbf{GAMa-Net} & \checkmark & 15.2 & 27.2 & \textbf{33.8} & \textbf{91.9} \\
\textbf{GAMa-Net} (Hierarchical) & \checkmark & \textbf{18.3} & \textbf{27.6} & 32.7 & - \\ 
\hline
\end{tabular}
\end{center}
\end{table}
\setlength{\tabcolsep}{1.4pt}
\vspace{-20pt}

\subsection{Results}

The evaluations of GAMa-Net and a comparison with baselines is shown in Table \ref{table:method_comparison}. 
We observe a Top-1 recall rate of 18.3\% and 15.2\%, using GAMa-Net with and without the hierarchical approach, respectively. On UCN set, we observe poor performance as compared to CN set which was expected. 

\textit{\textbf{Comparison:}}
In the first row, we show results with our image-based baselines which uses a single frame from the ground video. We also compare with other image based methods using a single frame as input and observe that the proposed approach outperforms all these baselines. While training with a single image is faster we do not have the temporal information which in this case can be perceived as relative positioning or contextual information. As the camera moves along a path or trajectory we can see the buildings or objects pass-by giving an idea of their respective location. The information of distance/relative-positioning as seen by a 3D-CNN is the additional information when we train with videos.  As shown in the Table \ref{table:method_comparison}, using a video provides better results as compared to images. 


Few studies have reported better performance with contrastive loss \cite{radenovic2018fine}, however triplet loss is also frequently used for geolocalization \cite{zhu2021vigor,regmi2021video}. 
We also observe better performance with contrastive loss as compared to the triplet loss (Table \ref{table:method_comparison}). 
In GAMa-Net, we have features from two different visual modalities i.e. one from ground \textbf{videos} and other from aerial \textbf{images}, which is different from the traditional contrastive loss. Similarly, training with triplet loss also use image-video features. These differences are likely responsible for poor training with triplet loss.

\begin{figure}[t!]
\begin{center}
\includegraphics[width=0.99\linewidth]{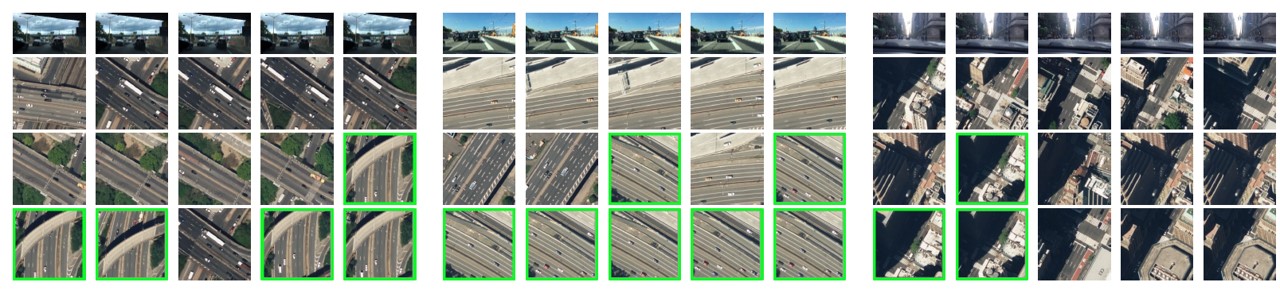}
\end{center}
  \caption{Three sample predictions by different models. Top-row shows frames of query clips. Second row is for combined model, and third row is GAMa-Net. Bottom row shows predictions by GAMa-Net with Hierarchical approach ( gallery reduced to 1\% of larger aerial regions). Correct predictions have a \textcolor{green}{green} outline. Please note that in case of close GPS labels there are multiple correct aerial images
  }
\label{fig:results_qualitative_1}
\end{figure}

\textit{\textbf{Qualitative analysis:}}
Figure \ref{fig:results_qualitative_1} show sample Top-5 predictions with different models where the leftmost is Top-1 and the rightmost is 5th. The combined model makes visually meaningful predictions. In the leftmost example, the camera is passing under a fly-over and the predictions show a similar locations. The middle example is of a road without any crossings or red lights in sight, and right-most example is of a city street with crossing markings on road. The predictions by combined model match these specifications. However, in these samples, the ground truth is in top-1\% but not in top-5 images. The predictions by GAMa-Net, with multi-headed self-attention improves the network performance and correct prediction moves up in the top-5 
(More results in supplementary).

\subsubsection{Hierarchical approach}

\setlength{\tabcolsep}{4pt}
\begin{table}[t!]
\begin{center}
\caption{\textcolor{blue}{Screening network} matches large aerial regions using a sequence of matched aerial images. In this Table, we compare 3D-CNN and 2D-CNN backbone for aerial image sequence of 8 or 32 images}
\label{screening_results}
\begin{tabular}{lcccc}
\hline\noalign{\smallskip}
Model & R@1 & R@10 & R@1\% & R@10\% \\
\noalign{\smallskip}
\hline
\noalign{\smallskip}
3D-CNN (seq of 8 aerial) & 7.5 & 24.9 & 39.8 & 67.9 \\ 
3D-CNN (seq of 32 aerial) & 3.8 & 19.8 & 36.1 & 74.9 \\ 
2D-CNN (seq of 8 aerial) & \textbf{12.2} & \textbf{35.3} & \textbf{49.3} & 77.2 \\ 
\textcolor{blue}{2D-CNN (seq of 32 aerial)} & 8.2 & 20.9 & 29.3 & \textbf{83.8} \\ 
\hline
\end{tabular}
\end{center}
\end{table}
\setlength{\tabcolsep}{1.4pt}

The predictions from the screening network are used for video-level geo-localization and to reduce the gallery for GAMa-Net. 
We experimented with both 2D-CNN and 3D-CNN backbone for aerial image sequence. As shown in Table \ref{screening_results}, better results were achieved with 2D-CNN backbone likely due to better spatial features. A comparison of number of input aerial images show that 2D-CNN with 8 frames retains more ground truths at Top-1 (12.2\%), Top-10 (35.3\%) and Top-1\% (49.3\%). We do not have 32 samples in some of the samples which likely effects the performance. However, when we evaluated GAMa-Net after gallery reduction, Screening network i.e. \textbf{2D-CNN network with 32 aerial images} provided the best results. 
Figure \ref{fig:network_screening}\textit{R} shows qualitative results for video-level geo-localizaion. It shows one frame of the video input to GAMa-Net, the predicted seq of aerial images thus obtained was used to identify the larger aerial regions. The bottom row shows the correct matched large aerial regions for video-level geo-localization. More results are in supplementary.

\setlength{\tabcolsep}{4pt}
\begin{table}[t!]
\begin{center}
\caption{Results with \textbf{GAMa-Net}, without and with hierarchical approach}
\label{table:hier_results}
\begin{tabular}{lcccc}
\hline\noalign{\smallskip}
\textbf{Gallery size} & R@1 & R@5 & R@10 & R@1\% \\
\noalign{\smallskip}
\hline
\noalign{\smallskip}
Full & 15.2 & 27.2 & 33.8 & \textbf{91.9} \\
\midrule
 & \multicolumn{4}{c}{\textbf{With Hierarchical Approach}} \\
\midrule
Top-10 & 16.2 & 22.8 & 25.5 & - \\ 
Top-1\% & \textbf{18.3} & 27.6 & 32.7 & - \\ 
Top-10\% & 16.6 & \textbf{28.2} & \textbf{34.6} & 76.6\\ 
\hline
\end{tabular}
\end{center}
\end{table}
\setlength{\tabcolsep}{1.4pt}

\textit{\textbf{Gallery sizes:}} In Table \ref{table:hier_results}, we discuss the results with GAMa-Net using various gallery sizes, i.e. Top-10, Top-1\%, and Top-10\% of large aerial regions, identified using screening network. We see better Top-1 results with Top-1\% gallery as compared to Top-10 and Top-10\%. There is a trade-off between reduced search space which improves matching and retaining the ground truth. As observed from Table \ref{screening_results}, in Top-10 we have ground truth for only 20.9\% of videos which increases to 83.8\% at 10\% gallery. However, this percentage will be different if we consider clips since we get different number of clips from each video, as per GPS labels. We evaluated GAMa-Net, which is clip level, using a reduced gallery. Even when ground truth was not available for many samples we see a Top-10 recall rate of 25.5\%.  
When gallery is reduced to Top-10 and Top-1\% of larger aerial regions we do not have enough clips to make upto 1\% of the total gallery.
In Figure \ref{fig:results_qualitative_1}, we show three examples of aerial image matching and it is evident that predictions by GAMa-Net are improved with hierarchical approach. 

\setlength{\tabcolsep}{1pt}
\begin{table}[t!]
\begin{center}
\caption{Results with our best model and ablations for comparison. Here we show Top-1@threshold for comparison, with threshold values of 0.1, 0.2, 0.5 and 1.0 mile. CBn-UCN and CBn-CN, represents contrastive baselines with uncentered and centered aerial images, respectively. Hierarchical approach uses one percent of the gallery}
\label{table:ablations_metric}
\begin{tabular}{lcccccccc}
\hline\noalign{\smallskip}
Model & \multicolumn{8}{c}{Recall @} \\
 & Top-1 & Top-5 & Top-10 & Top-1\% & Top-1@0.1 & Top-1@0.2 & Top-1@0.5 & Top-1@1.0\\
\noalign{\smallskip}
\hline
\noalign{\smallskip}
CBn-UCN & 3.2 & 10.9 & 16.7 & 65.0 & 3.6 & 5.9 & 11.3 & 18.6 \\
CBn-CN & 11.5 & 21.7 & 27.8 & 89.2& 15.7 & 19.0 & 25.1 & 32.1\\
Combined & 11.6 & 23.4 & 30.4 & \textbf{92.4} & 15.6 & 19.0 & 25.0 & 32.7 \\
\midrule
Video-Tx & 14.1 & 25.4 & 31.8 & 90.8 & 18.7 & 22.1 & 28.1 & 35.7 \\
GAMa-Net & 15.2 & 27.2 & \textbf{33.8} & 91.9 & 19.6 & 23.0 & 28.7 & 36.1\\
Dual-Tx & 14.6 & 26.1 & 32.7 & 91.8 & 19.0 & 22.2 & 28.2 & 35.5\\
%
\midrule
Hierarchical & \textbf{18.3} & \textbf{27.6} & 32.7 & - & \textbf{23.5} & \textbf{27.8} & \textbf{34.9} & \textbf{43.6}\\ 
\hline
\end{tabular}
\end{center}
\end{table}
\setlength{\tabcolsep}{1.4pt}


\subsection{Ablations} 
\textit{\textbf{Combined model:}}  In Table \ref{table:ablations_metric}, we observe better performance with combined model as compared to baselines, and a Top-1\% recall of 92.4\%. As expected, the performance improves as we increase the distance threshold. In the combined model, we have various augmentations which includes spatial and temporal centering, and random crop. Since images are not aligned in GAMa we stochastic-ally rotate the aerial view (0, 90, 180, 270 deg) for view-point invariance. All these augmentations have been reported to help improve geo-localization. We also include hard negatives for better training however transformer encoder is not a part of combined model. 

\textit{\textbf{Transformer encoder:}} 
Similar to an aerial image, not all visual features in the ground video have the same importance for cross-view geolocalization. Thus, we implemented transformer encoder on ground video(Video-Tx) and aerial images(Aerial-Tx), individually. In both cases, we observe an increase in recall @Top-1. The performance is better with GAMa-Net(i.e. Aerial-Tx) and we observe 15.2\% Top-1 recall which is higher than 14.1\% with Video-Tx (Table \ref{table:ablations_metric}). Observing an improvement in both cases we used transformer encoder on both sides(Dual-Tx) however it did not perform better than the GAMa-Net. 


\textit{\textbf{Hierarchical approach:}} We use hierarchical approach to improve the performance of GAMa-Net by reducing the gallery. Top-1@threshold (Table \ref{table:ablations_metric}) shows that using the hierarchical approach makes the matching more effective by predicting the aerial images closer to the correct GPS location or ground truth. The \textbf{difference} in Top-1 recall, with and without hierarchical approach, is even higher at higher thresholds i.e 7.5\% at 1.0 mile vs 3.9\% at 0.1 mile. An increase in recall to 43.6\% at 1.0 mile threshold shows that the ground truth is not very far from the Top-1.

\section{Discussion}

\textit{\textbf{Videos and contextual information:}} In videos, we have more information which can be considered as contextual information from geo-localization point of view. It enables the network to locate a given frame or view with respect to the other frames in the video. One frame may contain features to complement another and together they are likely to provide more or complete information required for geo-locating. In Table \ref{table:method_comparison}, we have compared video based method with frame based baselines. Better recall rate with videos show that the network is able to utilize the additional information available with videos. 

\textit{\textbf{Centered vs Uncentered:}} 
In an UNC aerial image the corresponding GPS point can be anywhere in the tile. In cases where the GPS point is near the boundary, the visual information from the video is less likely to match the corresponding tile. 
As expected, after centering (CBn-CN) we observed an improvement in the recall rate as compared with the uncentered set (i.e. CBn-UCN). 

\setlength{\tabcolsep}{4pt}
\begin{table}[t!]
\begin{center}
\caption{Training on full dataset and evaluation on day videos}
\label{results_full_dataset}
\begin{tabular}{lcccc}
\hline\noalign{\smallskip}
Model & \multicolumn{4}{c}{Recall @} \\
 & Top-1 & Top-1\% & Top-1@0.1 & Top-1@1.0  \\
\noalign{\smallskip}
\hline
\noalign{\smallskip}
Combined & 14.7 & \textbf{94.8} & 19.2 & 36.7 \\
GAMa-Net & 17.5 & 94.7 & 22.3 & 39.3 \\
Hierarchical & \textbf{19.4} & - & \textbf{25.0} & \textbf{45.1} \\
\hline
\end{tabular}
\end{center}
\end{table}
\setlength{\tabcolsep}{1.4pt}

\textit{\textbf{Full Dataset:}}
GAMa dataset is a large dataset which has its pros and cons. With large amount of data networks are better trained however this also increases the training time and memory requirement. Here we discuss results with models trained on full dataset (Table \ref{results_full_dataset}) i.e. both day and night videos. There is an improvement of 1-4\% in recall at all k. Thus, using the hierarchical approach our best R@Top-1 and R@Top-1@1.0 is 19.4\% and 45.1\%, respectively.  

\textit{\textbf{Comparison:}} Our results are comparable to existing image based methods. One recent study reports \textbf{R@1=13.95\% on CVUSA} when images are unaligned and have around 70\% field of view (FOV) \cite{yang2021cross}. Most of the cross-view geolocalization datasets such as CVUSA report high R@k while using ground panorama and aligned aerial images which is unrealistic since a normal camera lens has a FOV of around 72\% and it is not possible to get aligned images always. In GAMa dataset, the aerial images and ground videos are \textbf{unaligned}, and ground videos have a \textbf{limited FOV} which makes it a more realistic and difficult dataset.

\textit{\textbf{Challenges:}} In GAMa dataset among the two sets of small \textbf{aerial images}, i.e. CN and UCN, the UCN set is more realistic however difficult for geolocalization. Unaligned aerial images increase this difficulty however, the orientation information can be extracted from the GPS information and is likely to improve the performance \cite{liu2019lending}. Also, the \textbf{ground videos} have varying lengths and in some of the cases GPS label is not available every second. Thus, the video length available for geolocalization is less than 40 sec. Additionally, there is occlusion because of the car hood and other objects. Such cases are more likely to appear in fail cases. 

\textit{\textbf{Limitations:}} The proposed hierarchical approach performs better with longer videos (8 sec. or more), however it can be used with shorter clips as well. The aim behind using a hierarchical approach is to filter out confusing samples to improve the retrieval rate. However, this sometimes leads to filtering of the ground truth from the gallery. The model is also likely to fail with indoor videos since it will not be possible to match the features with an aerial image. However, this limitation is common to all cross-view geo-localization methods.

\section{Conclusions}

In this work, we focus on the problem of video geolocalization via cross-view matching. We propose a new dataset for this problem which has more than 51K ground videos and 1.9 million satellite images. The dataset spans multiple cities and is a more realistic dataset, unaligned and limited FOV. 
We believe that this dataset will be useful for future research in cross-view video geo-localization. Our proposed GAMa-Net effectively makes use of the rich contextual information available with video. In addition, we propose a hierarchical approach which also utilize the contextual information to further improve the geo-localization.

\clearpage
%
%
\bibliographystyle{splncs04}
\bibliography{main2022}
\clearpage

\appendix

\section{Overview}

In this supplementary, we have included additional qualitative results.
In Section \ref{qualitative_geo}, we show and discuss some qualitative results for video-level geolocalization. In Section \ref{additional_qualitative}, we have included additional qualitative results for clip-level geolocalization and ablations. 

\begin{figure}
\begin{center}
\includegraphics[width=0.6\linewidth]{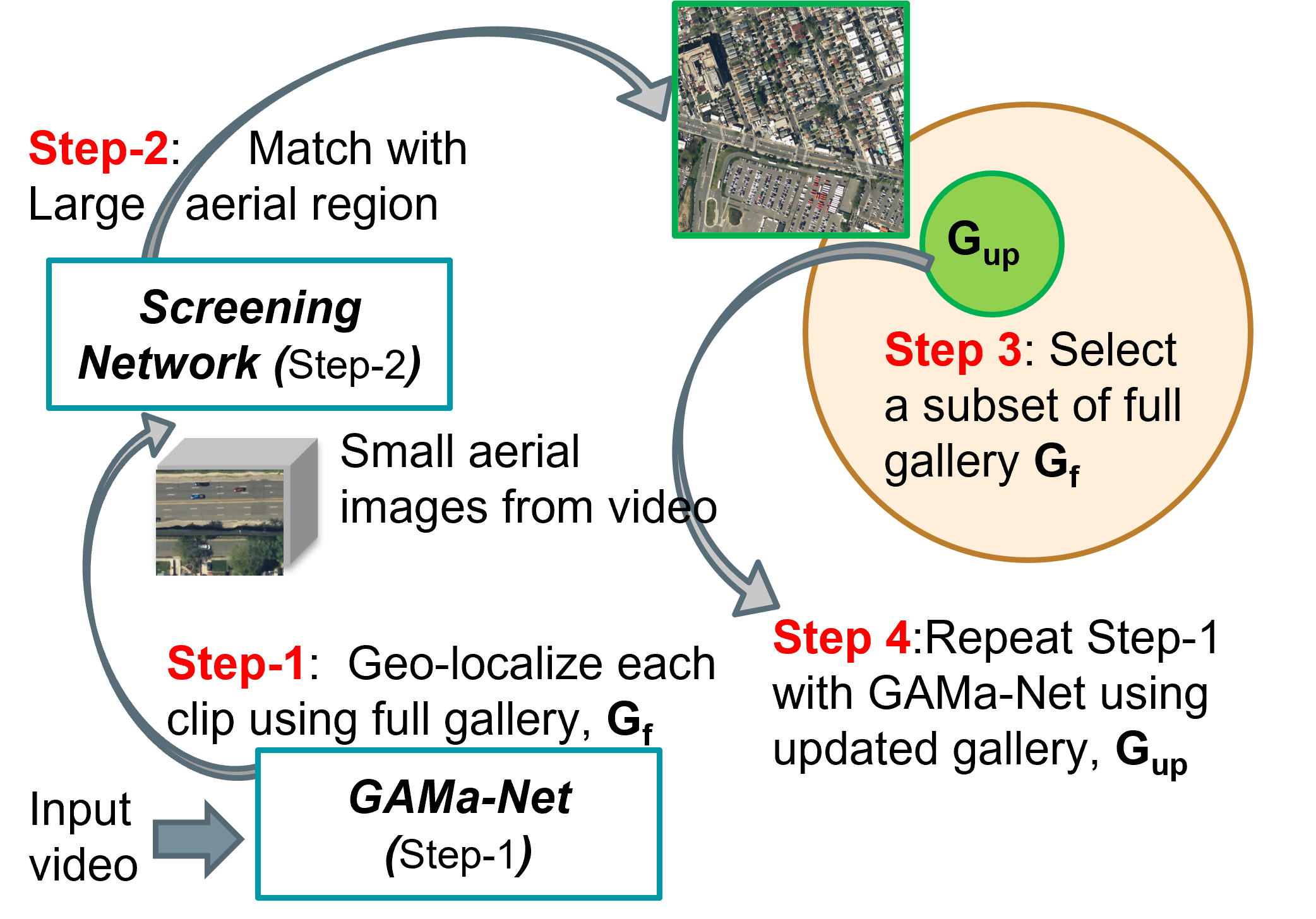}
\end{center}
  \caption{An outline of the proposed approach. Given a ground video, in Step-1 0.5 sec clips from the video are input to GAMa-Net. It takes one clip at a time and matches it with an aerial image. In Step-2, the sequence of aerial images obtained from GAMa-Net is input to the Screening network to retrieve the corresponding larger aerial region. This is our video-level geo-localization. In Step-3, top predictions of these larger aerial regions provides the updated gallery for a video. Step-4, is prediction by GAMa-Net while using the updated gallery}
\label{fig:supp_approach}
\vspace{-10pt}
\end{figure}

An \textbf{outline} of the proposed approach in Figure \ref{fig:supp_approach} shows how video level geo-localization is used to improve the clip-level results. The GAMa-Net outputs aerial image predictions at clip-level using the full gallery i.e. $G_{f}$. A video comprises of a number of clips (upto 40 clips per video), thus a sequence of aerial images is obtained from each query video, this sequence is then input to the screening network. The screening network uses this sequence of small aerial images to predict a large aerial region corresponding to the query video. Then we select Top-1\% large aerial images to update the gallery (the updated gallery $G_{up}$) for GAMa-Net and reduce the search space. We are hopeful that with further research this approach has the potential to be generalized to even larger scale. As shown in main paper we observe an improvement in Top-1 recall using this approach. The screening of the locations at large region level removed the confusing samples.

\section{Qualitative Results: \textit{Video-level Geo-localization}}
\label{qualitative_geo}
\begin{figure*}
\begin{center}
\includegraphics[width=0.98\linewidth]{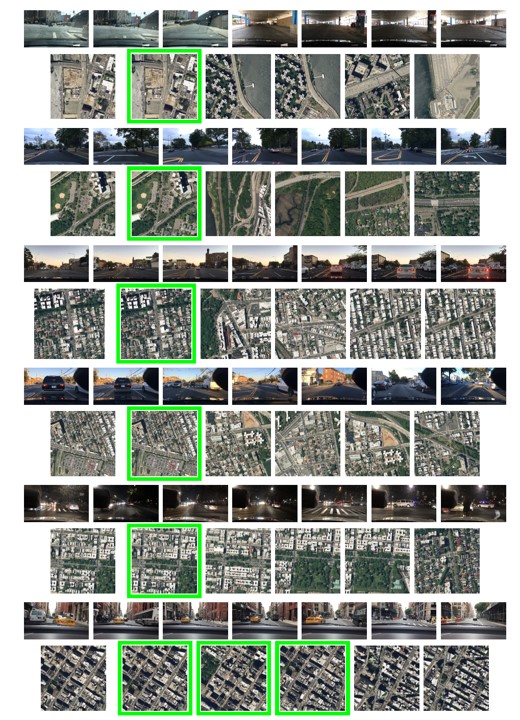}
\end{center}
\vspace{-15pt}
  \caption{Here, we show results for video-level geo-localization using the Screening network. In each sample, we show seven frames of the query video, followed by the ground truth aerial image and top-5 predictions of larger aerial regions.}
\label{fig:results_qualitative_screening}
\end{figure*}

In Figure \ref{fig:results_qualitative_screening}, we show some of the predictions of larger aerial regions as ranked by the screening network. For each sample, the first row shows frames of the video spanning over 35 sec. of time duration. In the subsequent row, we have the ground truth larger aerial region which is followed by Top-5 predictions by the screening network. The examples shown here are mostly correct predictions where Top-1 is same as the ground truth.

In the first sample (Figure \ref{fig:results_qualitative_screening}), we see that the video frames are \textbf{partially outdoors and partially indoors}. Indoor setting appears to be a parking lot or an underpass. The screening network is able to correctly localize the larger aerial region while using the information available from all the clips. The prediction when used to reduce the gallery of GAMa-Net is likely to enable a better prediction by screening-out the far away regions. Frame-by-frame it would have been difficult to localize the indoor frames. However, using the hierarchical approach the network is able to use the context from outdoor frames. We also observed this from the predictions, with GAMa-Net only 12 clips out of the 38 clips had ground truth prediction in top-10 and after reducing the gallery using the prediction from screening network this number increased to 28 clips. In the second sample (Figure \ref{fig:results_qualitative_screening}), we can see that all Top-5 predictions are visually similar. These predictions appears to be from the same region and most are from around a mile radius of the ground truth larger aerial region. 

In the third sample, because of a car, there is \textbf{occlusion} in part of the video. GAMa-Net correctly localizes the initial clips however fails in the clips with occlusion. After screening the gallery using the correct larger aerial prediction, most of these clips were correctly geolocalized by GAMa-Net in Top-1, Top-5 or Top-10. Similarly, in fourth and fifth sample the occlusion is in all the frames either because of the car hood or rear view mirror. In the fourth sample, similar improvement in clip-level geolocalization was observed with GAMa-Net because of correct screening of the larger aerial region at the video-level. We see correct video-level Top-1 with fifth sample, however the improvement in the clip-level predictions by GAMa-Net was less and only 3-4 more clips had predictions added to top-10 as compared to retrieval from the full gallery. The second top prediction of this sample is visually similar to the ground truth however a closer look shows that it is a different image. In the last or sixth sample we can see that multiple correct predictions because of the visual similarity are in Top-5.

With all these video samples, because of some of the correct clip-level predictions by GAMa-Net, the screening network is able to localize at video-level and identify the correct larger aerial region. However, the last sample had correct clip-level prediction for a single clip 
out of 31. It is likely that the \textbf{visual similarity} of the incorrect aerial image predictions helped with correct video-level geolocalization at Top-1 using screening network. 

\section{Additional Qualitative Results: \textit{Clip-level Geo-localization}}
\label{additional_qualitative}

\begin{figure*}[h]
\begin{center}
\includegraphics[width=0.98\linewidth]{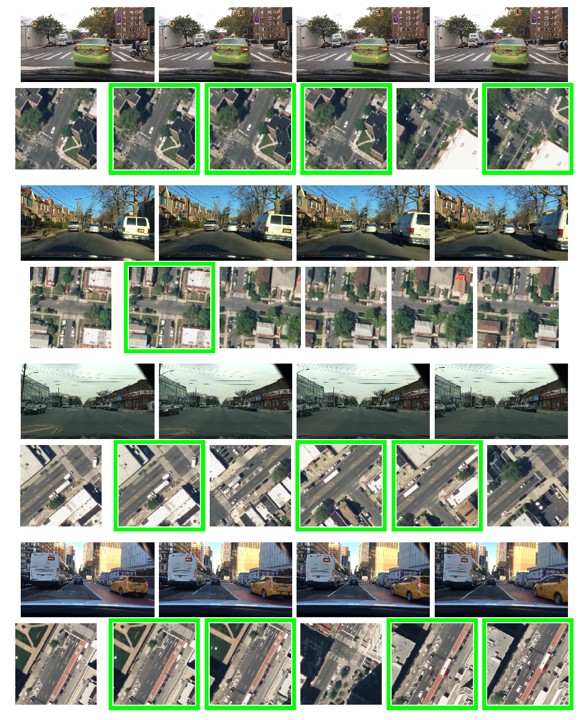}
\end{center}
  \caption{Here we show sample \textit{success cases} with GAMa-Net using Hierarchical approach where Top-1 prediction is correct. In each sample, four frames of the query clips are followed by ground truth aerial image and Top-5 predictions}
\label{fig:results_qualitative_hie}
\end{figure*}
\begin{figure*}[h]
\begin{center}
\includegraphics[width=0.98\linewidth]{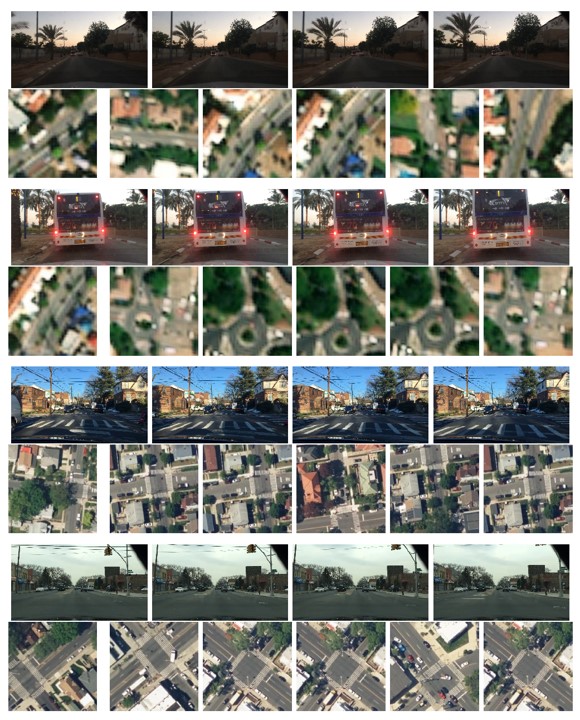}
\end{center}
  \caption{Here we show sample \textit{fail cases} with GAMa-Net using Hierarchical approach where Top-1 prediction is correct. In each sample, four frames of the query clips are followed by ground truth aerial image and Top-5 predictions}
\label{fig:results_qualitative_hie_fail}
\end{figure*}

In Figure \ref{fig:results_qualitative_hie} and Figure \ref{fig:results_qualitative_hie_fail}, we show additional results using the proposed GAMa-Net with Hierarchical approach, where we use the video-level predictions to improve clip-level geo-localization. Figure \ref{fig:results_qualitative_hie} shows examples of correct Top-1 predictions. We can see that most of the Top-5 predictions are nearby the ground truth. Because of the nearby GPS labels in video clips we have overlapping in aerial images and all these images appear in the Top-5 predictions along with the ground truth. Figure \ref{fig:results_qualitative_hie_fail} shows examples of fail cases, we can see that most of the fails are due to shadows or occlusion or poor quality of aerial images. The model makes meaningful predictions however fails in difficult or confusing samples e.g. in the last sample traffic lights are visible in the video and predictions are with zebra crossings however does not retrieve the correct aerial image in Top-5.

\begin{figure*}[h]
\begin{center}
\includegraphics[width=0.98\linewidth]{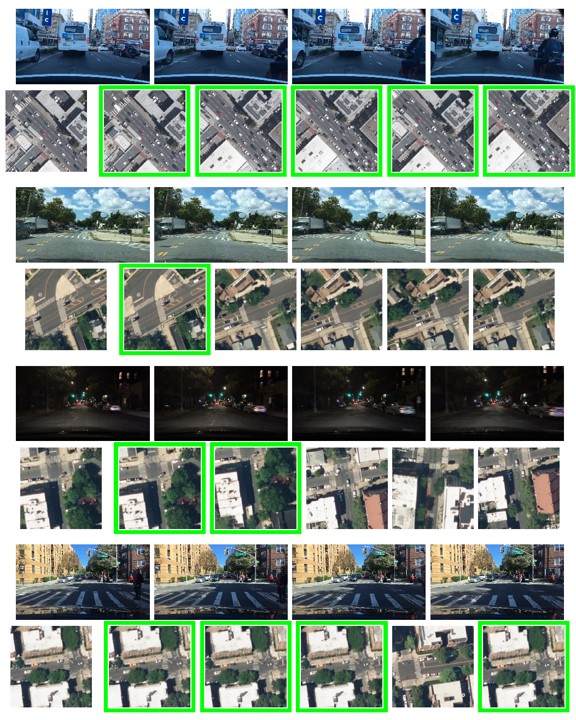}
\end{center}
  \caption{Here we show \textit{success cases} using GAMa-Net without Hierarchical approach where top-1 prediction is correct. In each sample, four frames of the query clips are followed by the ground truth aerial image and Top-5 predictions}
\label{fig:results_qualitative_gama}
\end{figure*}

\begin{figure*}
\begin{center}
\includegraphics[width=0.98\linewidth]{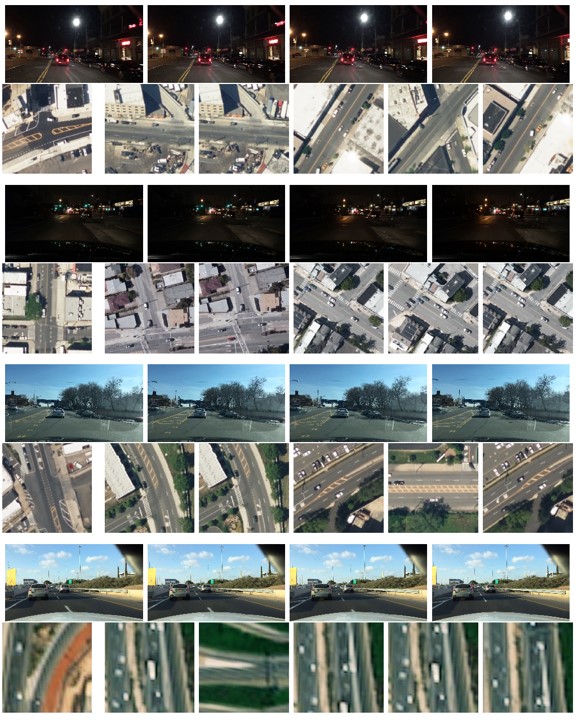}
\end{center}
\caption{Here we show \textit{fail cases} using GAMa-Net without Hierarchical approach. In each sample, four frames of the query clips are followed by the ground truth aerial image and top-5 predictions}
\label{fig:results_qualitative_gama_fail}
\end{figure*}

\begin{figure*}
\begin{center}
\includegraphics[width=0.98\linewidth]{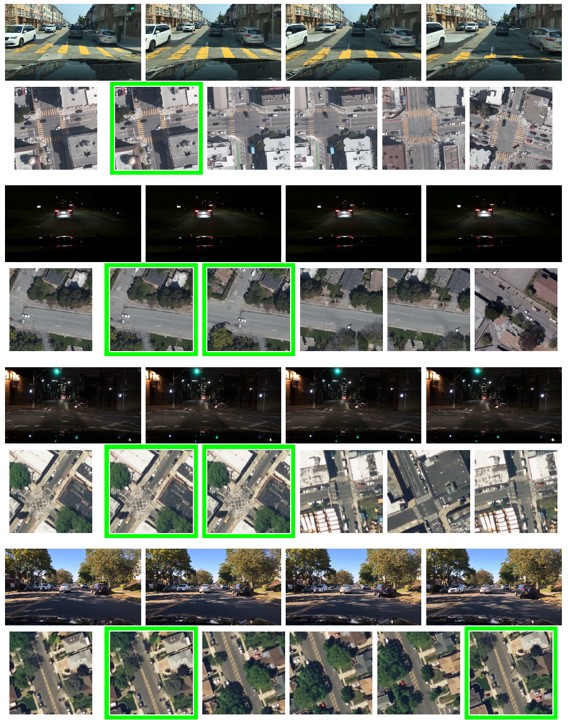}
\end{center}
\caption{Here we show \textit{success cases} for combined model where Top-1 prediction is correct. In each sample, four frames of the query clips are followed by ground truth aerial image and top-5 predictions}
\label{fig:results_qualitative_combined}
\end{figure*}
\begin{figure*}
\begin{center}
\includegraphics[width=0.98\linewidth]{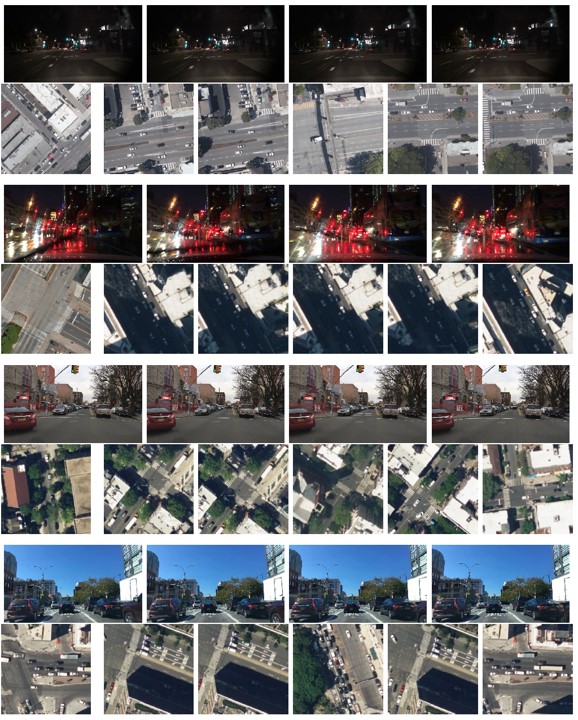}
\end{center}
\caption{Here we show \textit{fail cases} for combined model. In each sample, four frames of the query clips are followed by ground truth aerial image and Top-5 predictions}
\label{fig:results_qualitative_combined_fail}
\end{figure*}

In Figure \ref{fig:results_qualitative_combined} and Figure \ref{fig:results_qualitative_combined_fail}, we show some additional results with the combined model which does clip-level geolocalization by retrieving a matching aerial image. The network however is an ablation of the proposed GAMa-Net and does not have a transformer encoder.

\end{document}